\documentclass[10pt,twocolumn,letterpaper]{article}

\usepackage{cvpr}
\usepackage{times}
\usepackage{epsfig}
\usepackage{graphicx}
\usepackage{amsmath}
\usepackage{amssymb}

\usepackage[breaklinks=true,bookmarks=false]{hyperref}

\cvprfinalcopy 

\def\cvprPaperID{****} 

\begin{document}

\title{Deep Learned Full-3D Object Completion from Single View}

\author{D. Rethage$^1$, F. Tombari$^{1,2}$, F. Achilles$^1$, N. Navab$^{1,3}$\\
CAMP, Technische Universit\"at M\"unchen, Germany\\
DISI, University of Bologna, Italy\\
CAMP, Johns Hopkins University, USA\\
{\tt\small dario@rethage.net, \{tombari,achilles,navab\}@in.tum.de }
}

\maketitle

\begin{abstract}
   3D geometry is a very informative cue when interacting with and navigating an environment. This writing proposes a new approach to 3D reconstruction and scene understanding, which implicitly learns 3D geometry from depth maps pairing a deep convolutional neural network architecture with an auto-encoder. A data set of synthetic depth views and voxelized 3D representations is built based on ModelNet, a large-scale collection of CAD models, to train networks. The proposed method offers a significant advantage over current, explicit reconstruction methods in that it learns key geometric features offline and makes use of those to predict the most probable reconstruction of an unseen object. The relatively small network, consisting of roughly 4 million weights, achieves a 92.9\% reconstruction accuracy at a 30x30x30 resolution through the use of a pre-trained decompression layer. 
This is roughly 1/4 the weights of the current leading network.
The fast execution time of the model makes it suitable for real-time applications.
\end{abstract}

\section{Introduction}

An ongoing research interest in the area of scene understanding and robotic perception is the minimization of viewpoints necessary to reconstruct the 3D geometry of an object. Particularly, the maturation of mobile robots capable of interacting and navigating unfamiliar environments, and the recent availability of low-cost depth sensors puts emphasis on methods for reconstructing 3D geometry of objects in a robot's field-of-view. The proposed approach employs a deep convolutional neural network (CNN) to learn generic geometric features and makes use of these to carry out object completion by means of a single depth map. An auto-encoder is used to learn a compressed representation of object geometries in order to achieve high-resolution reconstruction while drastically reducing the number of parameters in the final layers of the network. The auto-encoder is separately trained on uniformly spaced sub-regions of voxelizations and is stacked on the end of the CNN for fine-tuning. This allows the CNN to, in effect, regress on a compressed representation of geometric feature voxelizations while still producing high resolution reconstructions.

\begin{figure}[t]
\begin{center}
\includegraphics[width=1\linewidth]{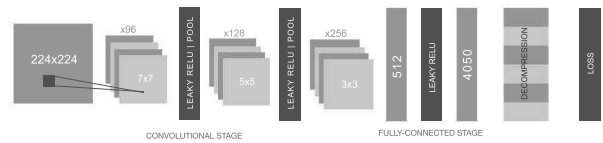}
\end{center}
   \caption{CNN + Decompressor Architecture}
\label{fig:long}
\label{fig:onecol}
\end{figure}

\subsection{Related Work}

Early works in the area of 3D reconstruction proposed sparse methods involving a minimum of two views, which attempt to explicitly calculate the depth of a set of points in a set of images. The 8-point algorithm\cite{Longuet-Higgins:1987} by Longuet-Higgins in 1981 followed by Hartley's normalized adaptation in 1997 make use of the epipolar constraint between a minimum of 8 points to determine corresponding depths. The two-view case was generalized by Weng, Liu and Huang \cite{WengLiuHuang:1988} as well as Spetsakis and Aloimonos \cite{SpetsakisAloimonos:1987} to support an arbitrary number of views with the introduction of the \lq trifocal tensor\rq, which captures the trilinear relationship of points between more than two views. Most recently in 2015, Wu et al. trained a convolutional deep belief network on depth maps and volumetric representations of objects to learn generic shape representations concurrently with object classifications \cite{WuSongKhoslaYuZhangTangXiao:2015}. As part of this work, the ModelNet data set was created, which is used throughout this project.

\section{Method}

Reconstruction at a 10x10x10 resolution was investigated first. This was followed by reconstruction of 30x30x30 voxelizations from 10 classes of ModelNet. Finally, the best performing high-resolution network was fine-tuned with a data set of 30 classes, in a sense applying curriculum learning \cite{bengio2009curriculum}.

\subsection{Architecture}

The CNN architecture consists of three convolutional layers and two fully-connected layers, with leaky RELU units used as non-linearities. The additional decompression layer, taken from the decompressing half of the auto-encoder, produces an overall six-layer architecture. Figure 1 depicts this architecture in detail. The reduction of weights is made possible by decompressing 27 uniformly spaced regions in parallel instead of the whole reconstruction at once. This technique is illustrated in figure 2. The result is an 85\% compression of the voxelization. While global relationships between different sub-regions within a voxelization are lost in the compressed representation, this information is captured in the previous layers of the CNN.

\subsection{Data set}

Networks were trained with three data sets. Each data set is created using a custom built rendering tool from a subset of ModelNet. One class is left out for testing. The 10-class data set consists of approximately 5000 models each with 8 depth viewpoints at 45 degree increments around the object resulting in almost 40,000 samples. The 30-class data set consists of more classes, but significantly fewer samples per class, namely, 100 models per class and only one viewpoint per model due to hardware limitations. To separately train an auto-encoder, the 10-class set is split into 27 uniformly split sub-grids.

\begin{figure}[t]
\begin{center}
\includegraphics[width=1\linewidth]{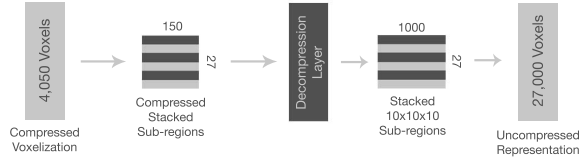}
\end{center}
   \caption{Compressing Auto-encoder}
\label{fig:long}
\label{fig:onecol}
\end{figure}

\subsection{Voxel Imbalance}

A central issue in training a neural network to regress on highly sparse representations is the risk of strong negative gradients killing hidden units in the network. This requires occupied and unoccupied voxels to be weighted differently. Unoccupied voxels are initially scaled down by a factor $s = \frac{\#\,\mathrm{occupied}}{\#\,\mathrm{unoccupied}}$ and gradually increase weight.

\section{Results}

\begin{figure}[t]
\begin{center}
\includegraphics[width=1\linewidth]{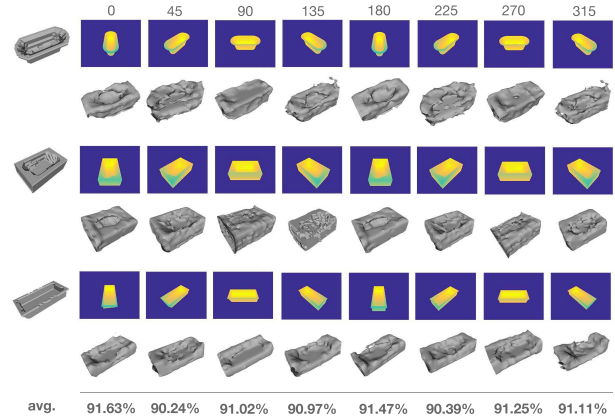}
\end{center}
   \caption{Examples from the unseen bathtub class}
\label{fig:long}
\label{fig:onecol}
\end{figure}

The low resolution network achieves an accuracy of 93\%, with no significant improvement produced from additional views. However, at this resolution many geometric features are not even captured in the ground-truth and therefore are impossible to learn in a network. The training of the best performing high-resolution network involved several manipulations at points where the network converged to a sub-optimal minima. Most importantly, changing unoccupied voxel weighting and allowing the decompression layer to update its weights only after several hundred epochs. Figure 3 depicts three instances of the test class from 8 view-points with corresponding reconstructions. The numbers in the bottom row represent the average reconstruction accuracy of the corresponding angle across all instances in the test set. The 10-class network achieves an accuracy of 91.01\% while the 30-class network improves to 92.9\%.

\section{Discussion and Conclusion}

This work has shown that it is possible to learn deep geometric representations based on a single depth view using a network of only 4 million weights. In addition, reconstruction accuracy is shown to be robust to the angle at which a depth map is captured. This research offers new insight into the challenges of learning highly sparse representations as well as how to compress geometric representations efficiently using a sub-region compression approach.

This work achieves similar reconstruction quality to 3D Shapenets\cite{WuSongKhoslaYuZhangTangXiao:2015} with 1/4 the amount of parameters. The network reconstructs based on one depth map in 27\,ms on a conventional CPU making it possible to reconstruct multiple objects in parallel in real-time scenarios. 
Moreover, the results of the 30-class fine-tuning show that it is possible to adapt a previously trained network on a significantly more diverse data set with a far smaller number of additional training samples.

The ability to efficiently carry out full-3D object completion from a single depth map enables new approaches in scene understanding, since the richer representation obtained by our network can be used to improve object retrieval, semantic inference and robotic manipulation.

\nocite{*}
{\small
\bibliographystyle{acm}
\bibliography{mybib}
}

\end{document}